\documentclass[10pt,a4paper]{article}%
\usepackage[T1]{fontenc}%
\usepackage[utf8]{inputenc}%
\usepackage{lmodern}%
\usepackage{textcomp}%
\usepackage{lastpage}%
\usepackage{times}%
\usepackage{hyperref}%
\usepackage{acl}%
\title{Sri Lanka Document Datasets: A Large{-}Scale, Multilingual Resource for Law, News, and Policy}%
\author{Nuwan I. Senaratna\\Independent Researcher\\\vspace{0.25em}\texttt{\href{https://github.com/nuuuwan}{https://github.com/nuuuwan}}}%
\date{\today}%
\begin{document}%
\normalsize%
\maketitle%
\begin{abstract}%
We present a collection of open, machine-readable document datasets covering parliamentary proceedings, legal judgments, government publications, news, and tourism statistics from Sri Lanka. The collection currently comprises of  299,131  documents (85.4 GB) across 26  datasets in Sinhala, Tamil, and English. The datasets are updated daily and mirrored on GitHub and Hugging Face. These resources aim to support research in computational linguistics, legal analytics, socio-political studies, and multilingual natural language processing. We describe the data sources, collection pipeline, formats, and potential use cases, while discussing licensing and ethical considerations. This manuscript is at version v2026-10-12-0839.%

\end{abstract}%
\section{Introduction}%
\label{sec:Introduction}%
Sri Lanka’s digital record of law, policy, and media is fragmented across numerous government and private sources. Much of this information exists as PDFs or web pages, often lacking machine- readable structure or public archival consistency. This fragmentation limits access for citizens, journalists, and researchers interested in the island’s governance, history, and socio-economic trends.%

The Sri Lanka Document Datasets initiative aims to bridge this gap by collecting, cleaning, and organizing key public documents into standardized, machine-readable formats. It unifies diverse materials—from Hansards and court judgements to news articles and tourism reports—under a common data framework. All datasets are openly licensed and continuously updated to ensure reproducibility and public transparency.%

This effort is particularly significant for data- driven research in low-resource contexts. By providing structured data in Sinhala, Tamil, and English, the project supports the development of natural language processing models, cross-lingual studies, and digital humanities research. The datasets also enable policy analysis, legal precedent tracking, and media monitoring in a transparent, open science environment.%

In this paper, we describe the scope and structure of these datasets, outline the scraping and curation processes, and highlight their potential applications in AI, governance, and public knowledge. Our goal is to create a living data archive that strengthens civic engagement and academic research through open, verifiable information.%

\section{Related Work}%
\label{sec:RelatedWork}%
The study of open datasets has been central to the
 development of natural language processing (NLP)
 and computational social science. Large corpora such
 as Common Crawl%
~%
\footnote{\href{https://commoncrawl.org/}{https://commoncrawl.org/}}%
, Wikipedia
 Dumps%
~%
\footnote{\href{https://dumps.wikimedia.org/}{https://dumps.wikimedia.org/}}%
, and OpenWebText%
~%
\citep{openwebtext2019}%
have powered models that generalize across domains. However, these resources are dominated by data from high-resource languages and global institutions.%

Regional initiatives have sought to address this
 imbalance by creating domain-specific collections.
 Examples include the Indian Kanoon legal corpus%
~%
\footnote{\href{https://indiankanoon.org/}{https://indiankanoon.org/}}%
, the OpenSubtitles
 multilingual dataset%
~%
\footnote{\href{https://opus.nlpl.eu/OpenSubtitles.php}{https://opus.nlpl.eu/OpenSubtitles.php}}%
, and the African News Corpus%
~%
\footnote{\href{https://data.africanlp.org}{https://data.africanlp.org}}%
. Such datasxets have improved representation for under-resourced languages and enabled comparative linguistic research.%

In South Asia, efforts remain scattered and often
 focus on individual media outlets or institutions.
 Sri Lanka, in particular, lacks consolidated and
 machine-readable documentation of its public
 records. Prior datasets were either limited in size,
 language coverage, or temporal continuity%
~%
\footnote{\href{https://github.com/sltalk}{https://github.com/sltalk}}%
~%
\footnote{\href{https://github.com/SriLankaNLP}{https://github.com/SriLankaNLP}}%
.%

The Sri Lanka Document Datasets aim to fill this gap by aggregating diverse sources—parliamentary debates, court judgements, gazettes, press releases, and news—into a unified, open, and multilingual repository. This complements global initiatives by providing a structured view of a unique national information ecosystem.%

\section{Datasets}%
\label{sec:Datasets}%
As of v2026-10-12-0839, Sri Lanka Document Datasets
  consists of 26 datasets
 and are publicly accessible on GitHub.%
~%
\footnote{\href{https://github.com/nuuuwan/lk\_datasets}{https://github.com/nuuuwan/lk\_datasets}}%

\begin{enumerate}%
\item%
\textbf{Hansard 2020S}: A Hansard is the official verbatim record of parliamentary debates, preserving lawmakers’ words and decisions for history, law, and public accountability.\textit{ 261 documents, 3.5 GB, from 2023{-}11{-}17 to 2026{-}09{-}24.} Source: \href{https://www.parliament.lk}{https://www.parliament.lk}. Dataset: \href{https://github.com/nuuuwan/lk\_hansard/tree/data\_lk\_hansard\_2020s/data/lk\_hansard\_2020s}{lk\_hansard\_2020s}.%
\item%
\textbf{Hansard 2010S}: A Hansard is the official verbatim record of parliamentary debates, preserving lawmakers’ words and decisions for history, law, and public accountability.\textit{ 10 documents, 153.2 MB, from 2019{-}09{-}05 to 2019{-}11{-}11.} Source: \href{https://www.parliament.lk}{https://www.parliament.lk}. Dataset: \href{https://github.com/nuuuwan/lk\_hansard/tree/data\_lk\_hansard\_2010s/data/lk\_hansard\_2010s}{lk\_hansard\_2010s}.%
\item%
\textbf{Hansard 2000S}: A Hansard is the official verbatim record of parliamentary debates, preserving lawmakers’ words and decisions for history, law, and public accountability.\textit{ 10 documents, 135.4 MB, from 2009{-}10{-}20 to 2009{-}12{-}08.} Source: \href{https://www.parliament.lk}{https://www.parliament.lk}. Dataset: \href{https://github.com/nuuuwan/lk\_hansard/tree/data\_lk\_hansard\_2000s/data/lk\_hansard\_2000s}{lk\_hansard\_2000s}.%
\item%
\textbf{Appeal Court Judgements}: A Court of Appeal judgment is a higher court ruling that reviews decisions of lower courts, shaping legal precedent and protecting citizens’ rights.\textit{ 18,551 documents, 14.0 GB, from 2010{-}10{-}16 to 2026{-}09{-}25.} Source: \href{https://courtofappeal.lk}{https://courtofappeal.lk}. Dataset: \href{https://github.com/nuuuwan/lk\_appeal\_court\_judgements/tree/data/data/lk\_appeal\_court\_judgements}{lk\_appeal\_court\_judgements}.%
\item%
\textbf{Supreme Court Judgements}: A Supreme Court judgment is a binding legal decision that interprets the Constitution and laws, shaping justice, governance, and citizens’ rights.\textit{ 2,729 documents, 1.7 GB, from 2009{-}01{-}27 to 2026{-}05{-}27.} Source: \href{https://supremecourt.lk}{https://supremecourt.lk}. Dataset: \href{https://github.com/nuuuwan/lk\_supreme\_court\_judgements/tree/data/data/lk\_supreme\_court\_judgements}{lk\_supreme\_court\_judgements}.%
\item%
\textbf{Police Press Releases}: A police press release is an official update from law enforcement on crimes, arrests, safety alerts, or public notices, ensuring transparency and public awareness.\textit{ 1,434 documents, 513.0 MB, from 2025{-}05{-}01 to 2026{-}08{-}08.} Source: \href{https://www.police.lk}{https://www.police.lk}. Dataset: \href{https://github.com/nuuuwan/lk\_police\_press\_releases/tree/data/data/lk\_police\_press\_releases}{lk\_police\_press\_releases}.%
\item%
\textbf{Acts}: A legal act is a law passed by Parliament that governs rights, duties, economy, and society, shaping daily life and national policy.\textit{ 3,979 documents, 6.8 GB, from 1981{-}01{-}22 to 2026{-}04{-}09.} Source: \href{https://documents.gov.lk}{https://documents.gov.lk}. Dataset: \href{https://github.com/nuuuwan/lk\_legal\_docs/tree/data\_lk\_acts/data/lk\_acts}{lk\_acts}.%
\item%
\textbf{Bills}: A Bill is a draft law proposed in Parliament. It becomes binding once passed and enacted, shaping governance, rights, and daily life in the country.\textit{ 4,236 documents, 2.0 GB, from 2010{-}05{-}10 to 2026{-}04{-}29.} Source: \href{https://documents.gov.lk}{https://documents.gov.lk}. Dataset: \href{https://github.com/nuuuwan/lk\_legal\_docs/tree/data\_lk\_bills/data/lk\_bills}{lk\_bills}.%
\item%
\textbf{Extraordinary Gazettes 2020S}: An Extraordinary Gazette is an official government publication used to announce urgent laws, regulations, or public notices with immediate effect.\textit{ 49,513 documents, 11.6 GB, from 2020{-}01{-}01 to 2026{-}04{-}28.} Source: \href{https://documents.gov.lk}{https://documents.gov.lk}. Dataset: \href{https://github.com/nuuuwan/lk\_legal\_docs/tree/data\_lk\_extraordinary\_gazettes\_2020s/data/lk\_extraordinary\_gazettes\_2020s}{lk\_extraordinary\_gazettes\_2020s}.%
\item%
\textbf{Extraordinary Gazettes 2010S}: An Extraordinary Gazette is an official government publication used to announce urgent laws, regulations, or public notices with immediate effect.\textit{ 56,379 documents, 17.4 GB, from 2010{-}01{-}01 to 2019{-}12{-}31.} Source: \href{https://documents.gov.lk}{https://documents.gov.lk}. Dataset: \href{https://github.com/nuuuwan/lk\_legal\_docs/tree/data\_lk\_extraordinary\_gazettes\_2010s/data/lk\_extraordinary\_gazettes\_2010s}{lk\_extraordinary\_gazettes\_2010s}.%
\item%
\textbf{Cabinet Decisions}: A Sri Lanka Cabinet Decision is an official policy or action agreed by the Cabinet of Ministers, shaping governance, law, and national development in the country.\textit{ 11,126 documents, 143.5 MB, from 2010{-}09{-}27 to 2026{-}09{-}28.} Source: \href{https://www.cabinetoffice.gov.lk}{https://www.cabinetoffice.gov.lk}. Dataset: \href{https://github.com/nuuuwan/lk\_cabinet\_decisions/tree/data/data/lk\_cabinet\_decisions}{lk\_cabinet\_decisions}.%
\item%
\textbf{Treasury Press Releases}: A Sri Lanka Treasury press release shares key govt financial updates—on budgets, debt, or policy—vital for transparency, guiding investors, citizens, and markets on the nation’s economic direction.\textit{ 178 documents, 179.7 MB, from 2015{-}09{-}08 to 2026{-}10{-}12.} Source: \href{https://www.treasury.gov.lk}{https://www.treasury.gov.lk}. Dataset: \href{https://github.com/nuuuwan/lk\_treasury/tree/data\_lk\_treasury\_press\_releases/data/lk\_treasury\_press\_releases}{lk\_treasury\_press\_releases}.%
\item%
\textbf{Pmd Press Releases}: A Sri Lanka Presidential Media Division press release shares official updates on national decisions, policies, or events. It’s vital as the authoritative source ensuring transparency and public awareness.\textit{ 2,182 documents, 55.9 MB, from 2024{-}09{-}23 to 2025{-}09{-}24.} Source: multiple sources. Dataset: \href{https://github.com/nuuuwan/lk\_pmd/tree/data\_lk\_pmd\_press\_releases/data/lk\_pmd\_press\_releases}{lk\_pmd\_press\_releases}.%
\item%
\textbf{News}: A collection of news documents.\textit{ 132,830 documents, 2.1 GB, from 2021{-}09{-}12 to 2026{-}10{-}12.} Source: multiple sources. Dataset: \href{https://github.com/nuuuwan/lk\_news/tree/data/data/lk\_news}{lk\_news}.%
\item%
\textbf{Tourism Weekly Reports}: Report on Weekly Tourist Arrivals to Sri Lanka.\textit{ 46 documents, 119.6 MB, from 2023{-}01{-}01 to 2026{-}10{-}01.} Source: \href{https://www.sltda.gov.lk}{https://www.sltda.gov.lk}. Dataset: \href{https://github.com/nuuuwan/lk\_tourism/tree/data\_lk\_tourism\_weekly\_reports/data/lk\_tourism\_weekly\_reports}{lk\_tourism\_weekly\_reports}.%
\item%
\textbf{Tourism Monthly Reports}: Report on Monthly Tourist Arrivals to Sri Lanka.\textit{ 140 documents, 368.8 MB, from 2015{-}01{-}01 to 2026{-}09{-}01.} Source: multiple sources. Dataset: \href{https://github.com/nuuuwan/lk\_tourism/tree/data\_lk\_tourism\_monthly\_reports/data/lk\_tourism\_monthly\_reports}{lk\_tourism\_monthly\_reports}.%
\item%
\textbf{Dmc Situation Reports}: Situation Report including information about Heavy Rain, Wind, Tree Falling, Lighting etc.\textit{ 4,707 documents, 3.2 GB, from 2018{-}01{-}02 to 2026{-}10{-}10.} Source: \href{https://www.dmc.gov.lk}{https://www.dmc.gov.lk}. Dataset: \href{https://github.com/nuuuwan/lk\_dmc/tree/data\_lk\_dmc\_situation\_reports/data/lk\_dmc\_situation\_reports}{lk\_dmc\_situation\_reports}.%
\item%
\textbf{Dmc Weather Forecasts}: Weather Forecasts for various places in Sri Lanka.\textit{ 6,206 documents, 7.6 GB, from 2023{-}03{-}26 to 2026{-}10{-}12.} Source: \href{https://www.dmc.gov.lk}{https://www.dmc.gov.lk}. Dataset: \href{https://github.com/nuuuwan/lk\_dmc/tree/data\_lk\_dmc\_weather\_forecasts/data/lk\_dmc\_weather\_forecasts}{lk\_dmc\_weather\_forecasts}.%
\item%
\textbf{Dmc River Water Level And Flood Warnings}: River Water Level and Flood Warnings for various places in Sri Lanka.\textit{ 873 documents, 326.4 MB, from 2025{-}06{-}10 to 2026{-}10{-}12.} Source: \href{https://www.dmc.gov.lk}{https://www.dmc.gov.lk}. Dataset: \href{https://github.com/nuuuwan/lk\_dmc/tree/data\_lk\_dmc\_river\_water\_level\_and\_flood\_warnings/data/lk\_dmc\_river\_water\_level\_and\_flood\_warnings}{lk\_dmc\_river\_water\_level\_and\_flood\_warnings}.%
\item%
\textbf{Dmc Landslide Warnings}: Landslide Warnings including early warnings, locations of potential risk, areas and places which need special attention, and automated landslide early warning map.\textit{ 719 documents, 542.9 MB, from 2019{-}09{-}26 to 2026{-}10{-}11.} Source: \href{https://www.dmc.gov.lk}{https://www.dmc.gov.lk}. Dataset: \href{https://github.com/nuuuwan/lk\_dmc/tree/data\_lk\_dmc\_landslide\_warnings/data/lk\_dmc\_landslide\_warnings}{lk\_dmc\_landslide\_warnings}.%
\item%
\textbf{Cbsl Annual Reports}: Annual Reports of the Central Bank of Sri Lanka (CBSL).It has been discountinued since 2023and replaced with swo separate reports,namely, the Annual Economic Reviewand Financial Statementsand Operations of the Central Bank.\textit{ 1,137 documents, 3.5 GB, from 1950{-}01{-}01 to 2023{-}01{-}01.} Source: \href{https://www.cbsl.gov.lk}{https://www.cbsl.gov.lk}. Dataset: \href{https://github.com/nuuuwan/cbsl/tree/data\_cbsl\_annual\_reports/data/cbsl\_annual\_reports}{cbsl\_annual\_reports}.%
\item%
\textbf{Fisheries Annual Statistics Reports}: Annual Fisheries Statistics Reports of the Ministry of Fisheries,Aquatic and Ocean Resources, Sri Lanka\textit{ 9 documents, 15.1 MB, from 2017{-}01{-}01 to 2024{-}01{-}01.} Source: \href{https://www.fisheries.gov.lk}{https://www.fisheries.gov.lk}. Dataset: \href{https://github.com/nuuuwan/lk\_fisheries/tree/data\_lk\_fisheries\_annual\_statistics\_reports/data/lk\_fisheries\_annual\_statistics\_reports}{lk\_fisheries\_annual\_statistics\_reports}.%
\item%
\textbf{Fisheries Monthly Fish Production Reports}: Monthly Fish Production Reports of the Ministry of Fisheries,Aquatic and Ocean Resources, Sri Lanka\textit{ 102 documents, 9.7 MB, from 2019{-}01{-}01 to 2025{-}12{-}01.} Source: \href{https://www.fisheries.gov.lk}{https://www.fisheries.gov.lk}. Dataset: \href{https://github.com/nuuuwan/lk\_fisheries/tree/data\_lk\_fisheries\_monthly\_fish\_production\_reports/data/lk\_fisheries\_monthly\_fish\_production\_reports}{lk\_fisheries\_monthly\_fish\_production\_reports}.%
\item%
\textbf{Fisheries Weekly Fish Prices Reports}: Weekly Fish Prices Reports of the Ministry of Fisheries,Aquatic and Ocean Resources, Sri Lanka\textit{ 277 documents, 23.9 MB, from 2019{-}01{-}01 to 2026{-}04{-}08.} Source: \href{https://www.fisheries.gov.lk}{https://www.fisheries.gov.lk}. Dataset: \href{https://github.com/nuuuwan/lk\_fisheries/tree/data\_lk\_fisheries\_weekly\_fish\_prices\_reports/data/lk\_fisheries\_weekly\_fish\_prices\_reports}{lk\_fisheries\_weekly\_fish\_prices\_reports}.%
\item%
\textbf{Fisheries Monthly Export Import Reports}: Monthly Fish Export and Import Reports of the Ministry of Fisheries,Aquatic and Ocean Resources, Sri Lanka\textit{ 71 documents, 55.7 MB, from 2019{-}01{-}01 to 2026{-}02{-}01.} Source: \href{https://www.fisheries.gov.lk}{https://www.fisheries.gov.lk}. Dataset: \href{https://github.com/nuuuwan/lk\_fisheries/tree/data\_lk\_fisheries\_monthly\_export\_import\_reports/data/lk\_fisheries\_monthly\_export\_import\_reports}{lk\_fisheries\_monthly\_export\_import\_reports}.%
\item%
\textbf{Edupub}: Educational Publications from the Educational Publications Department, Sri Lanka.\textit{ 1,426 documents, 9.4 GB, from 2025{-}01{-}01 to 2025{-}01{-}01.} Source: \href{http://www.edupub.gov.lk}{http://www.edupub.gov.lk}. Dataset: \href{https://github.com/nuuuwan/lk\_edupub/tree/data/data/lk\_edupub}{lk\_edupub}.%
\end{enumerate}

\section{Data Collection Pipeline}%
\label{sec:DataCollectionPipeline}%
Our pipeline is automated, reproducible, and
 resilient. It continuously discovers, ingests,
 parses, validates, and versions documents from
 official Sri Lankan sources%
~%
\citep{MLOpsSurvey2022}%
.%

GitHub Actions orchestrates the workflow.
 Cron jobs run several times per day, per
 dataset. A matrix strategy isolates each source,
 allowing independent retries without blocking
 others. Secrets manage tokens; caches speed I/O%
~%
\footnote{\href{https://docs.github.com/actions}{https://docs.github.com/actions}}%
.%

Each run is idempotent and incremental.
 Before crawling, we load a manifest of known
 items. New or changed items are detected by
 stable keys (URL + date) and content hashes.
 Only deltas are committed to the repository%
~%
\citep{ReproducibleResearch2017}%
.%

Crawling is implemented in Python with
 Selenium in headless Chromium.
 We wait for dynamic content via explicit
 conditions (e.g., presence of selectors),
 handle pagination, and capture canonical URLs%
~%
\footnote{\href{https://www.selenium.dev/documentation/}{https://www.selenium.dev/documentation/}}%
.%

Politeness is enforced. We respect robots.txt,
 throttle requests, randomize delays, and apply
 exponential backoff on transient failures.
 Errors are logged and surfaced in the Actions
 summary for rapid triage%
~%
\citep{WebCrawlingBestPractices2021}%
.%

Raw artifacts are preserved alongside parsed
 representations. For each document we store
 the fetched HTML or PDF, plus normalized JSON
 with metadata (title, date, source, language,
 hashes) to enable downstream reproducibility%
~%
\citep{DataVersioning2020}%
.%

PDF parsing uses PyMuPDF (also known as fitz). For each PDF,
 we extract text, metadata, and layout blocks,
 retain page boundaries, and normalize Unicode.
 When images contain embedded text, PyMuPDF’s
 text extraction captures vector text regions%
~%
\footnote{\href{https://pymupdf.readthedocs.io}{https://pymupdf.readthedocs.io}}%
.%

The parser records coordinates for blocks, allowing approximate structure recovery (sections, headings, tables) where present. Heuristics join hyphenated lines and preserve numbering and legal citations .%

Quality gates run in CI. Schemas are validated, required fields are enforced, and checksums guard against corruption. Unit tests cover fetching, parsing, and serialization, and fail the job on regressions .%

Historical coverage was built via a back-population pipeline. We iterate over archival indexes and date ranges, enqueue jobs in batches, checkpoint progress, and resume safely after interruptions .%

Transparency is prioritized. Run metadata, document counts, and last-updated timestamps are published to README badges. Commit messages summarize deltas, enabling clear, auditable provenance across releases .%

\section{Licensing and Access}%
\label{sec:LicensingandAccess}%
All datasets and code are openly available to
 the public. The repositories are hosted on
 GitHub under the MIT License, which permits
 reuse, modification, and redistribution with
 attribution to the original author%
~%
\footnote{\href{https://opensource.org/licenses/MIT}{https://opensource.org/licenses/MIT}}%
.%

This permissive model encourages transparency
 and collaboration. Researchers, developers, and
 institutions can integrate the datasets into
 their pipelines without restrictive terms or
 commercial barriers%
~%
\citep{OpenDataPractices2020}%
.%

Each dataset repository includes structured
 metadata, versioned releases, and README files
 with descriptive statistics and provenance.
 All assets are mirrored on Hugging Face%
~%
\footnote{\href{https://huggingface.co/nuuuwan/datasets}{https://huggingface.co/nuuuwan/datasets}}%
to ensure redundancy and faster global access .%

Public accessibility is a design principle.
 Automated GitHub Actions update metadata badges
 and commit summaries whenever new data are
 ingested. Users can clone, fork, or download
 any subset directly without authentication%
~%
\footnote{\href{https://docs.github.com/actions}{https://docs.github.com/actions}}%
.%

The open license facilitates reproducible
 science and supports downstream applications
 in natural language processing, digital
 governance, and policy research. By ensuring
 public access, the project aligns with FAIR
 principles—Findable, Accessible, Interoperable,
 and Reusable%
~%
\citep{FAIRPrinciples2016}%
.%

\section{Conclusion and Future Work}%
\label{sec:ConclusionandFutureWork}%
This project establishes an open, reproducible,
 and scalable foundation for Sri Lankan document
 datasets, spanning legal, governmental, and
 media sources. The pipeline integrates crawling,
 parsing, and versioning into a unified
 ecosystem for data-driven research%
~%
\citep{OpenDataPractices2020}%
.%

The datasets already serve as a valuable
 resource for natural language processing,
 computational law, and policy analysis. They
 enable quantitative and qualitative insights
 into governance, lawmaking, and civic
 communication over time%
~%
\citep{FAIRPrinciples2016}%
.%

Future development focuses on three priorities:%

\begin{enumerate}%

\item First, expanding coverage by adding new datasets from additional government agencies, media sources, and historical archives .%

\item Second, improving the linguistic accuracy of Sinhala and Tamil parsing, particularly for complex sentence structures and transliterated terms. Enhancements in tokenization, font handling, and multilingual embeddings are planned .%

\item Third, integrating OCR parsing for PDFs with unstructured or scanned content. We are experimenting with deep-learning-based OCR pipelines that combine layout recognition and language modeling to recover high-quality text from low-quality sources .%

\end{enumerate}%

Together, these directions will further improve coverage, data quality, and usability, ensuring that the Sri Lanka Datasets initiative remains a sustainable open infrastructure for the region’s digital and academic ecosystem .%

\bibliographystyle{acl_natbib}%
\bibliography{latex/lk_datasets}%
\end{document}